\documentclass[letterpaper, 10 pt, journal, twoside]{IEEEtran}
\usepackage{amsmath,amsfonts}
\usepackage{algorithm}
\usepackage{array}
\usepackage[caption=false,font=normalsize,labelfont=sf,textfont=sf]{subfig}
\usepackage{textcomp}
\usepackage{stfloats}
\usepackage{url}
\usepackage{verbatim}
\usepackage{graphicx}
\usepackage{cite}
\usepackage{multirow}
\usepackage{amsmath} 
\usepackage{amssymb}  
\usepackage{threeparttable}
\usepackage[draft]{hyperref}

\usepackage[noend]{algpseudocode}
\usepackage{tikz}

\usetikzlibrary{arrows}
\usetikzlibrary{shapes}

\newcommand*\circled[1]{\tikz[baseline=(char.base)]{
            \node[shape=circle,draw,inner sep=0.1pt] (char) {#1};}}

\usepackage[]{xcolor}
\definecolor{orange}{RGB}{255, 128, 0}
\definecolor{purple}{RGB}{102, 0, 204}
\definecolor{blue}{RGB}{0, 51, 102}
\definecolor{reds}{RGB}{102, 0,51}

\begin{document}

\title{Optimization-based Trajectory Tracking Approach for Multi-rotor Aerial Vehicles in Unknown Environments}


\author{Geesara Kulathunga$^{1}$, Hany Hamed$^{2}$,  Dmitry Devitt$^{2}$, Alexandr Klimchik$^{2}$%
\thanks{Manuscript received: September, 9, 2021; Revised  December, 3, 2021; Accepted January, 29, 2022.}
\thanks{This paper was recommended for publication by Editor Stephen J. Guy upon evaluation of the Associate Editor and Reviewers' comments.
 This work was supported by The Analytical Center for the Government of the Russian Federation (Agreement No. 70-2021-00143 dd. 01.11.2021, IGK 000000D730321P5Q0002).} 
\thanks{$^{1}$Artificial Intelligence Research Center, Innopolis University, Russia
        {\tt\footnotesize ggeesara@gmail.com}}%
\thanks{$^{2}$ Centre for Robotics and Mechatronics Components, Innopolis University, Russia
        {\tt\footnotesize h.hamed@innopolis.university}, {\tt\footnotesize d.devitt@innopolis.ru}, {\tt\footnotesize a.klimchik@innopolis.ru}}%
\thanks{Digital Object Identifier (DOI): see top of this page.}
}
\markboth{ IEEE ROBOTICS AND AUTOMATION LETTERS. PREPRINT VERSION. ACCEPTED JANUARY, 2022}%
{Kulathunga \MakeLowercase{\textit{et al.}}: Optimization-based Trajectory Tracker} 


\maketitle

\begin{abstract}
The goal of this paper is to develop a continuous optimization-based refinement of the reference trajectory to 'push it out' of the obstacle-occupied space in the global phase for Multi-rotor Aerial Vehicles in unknown environments. Our proposed approach comprises two planners: a global planner and a local planner. The global planner refines the initial reference trajectory when the trajectory goes either through an obstacle or near an obstacle and lets the local planner calculate a near-optimal control policy. 
The global planner comprises two convex programming approaches: the first one helps to refine the reference trajectory, and the second one helps to recover the reference trajectory if the first approach fails to refine. The global planner mainly focuses on real-time performance and obstacles avoidance, whereas the proposed formulation of the constrained nonlinear model predictive control-based local planner ensures safety, dynamic feasibility, and the reference trajectory tracking accuracy for low-speed maneuvers, provided that local and global planners have mean computation times 0.06s (15Hz) and 0.05s (20Hz), respectively, on an NVIDIA Jetson Xavier NX computer. The results of our experiment confirmed that, in cluttered environments, the proposed approach outperformed three other approaches: sampling-based pathfinding followed by trajectory generation, a local planner, and graph-based pathfinding followed by trajectory generation.  
\end{abstract}


\section{INTRODUCTION}
\IEEEPARstart{T}he reference trajectory tracking for multi-rotor aerial vehicles (MAVs) is used in various domains, e.g., cinematography, or landing on a moving platform. Even though many approaches have been proposed for tracking specified reference trajectories~\cite{baca2018model, 9513301, mechali2021observer}, it remains an open research problem due to several reasons: achieving real-time performance, avoiding close-in obstacles, adhering to different weather conditions, etc. Subsequently, generating a near-optimal control policy for maneuvering through a cluttered unknown environment is a rather challenging task when enforcing the dynamic feasibility and safety constraints in real-time. Model Predictive Control (MPC) is one of the promising techniques to address such challenging tasks. However, due to the computational aspects of MPC, it is difficult to achieve real-time performance in most situations using limited available resources~\cite{mpc_mc}. 
Such aspects are mainly because of the way the problem is formulated, e.g., as an NMPC (Nonlinear MPC), as an LMPC (Linear MPC), and the way constrains, e.g., obstacles and inputs, are handled. Moreover, the accuracy of the near-optimal policy generation of MPC depends on the sensing capabilities, e.g., FoV (Field of View), sensing distance, and the way surrounded free space and obstacles are represented. For instance, free space can be formed as a set of convex polyhedrons along the refined reference trajectory. Afterwards, incorporating a linear motion model, the problem can be formulated as convex rather than non-convex.

\begin{figure}[]
    \centering
    \includegraphics[width=1\linewidth]{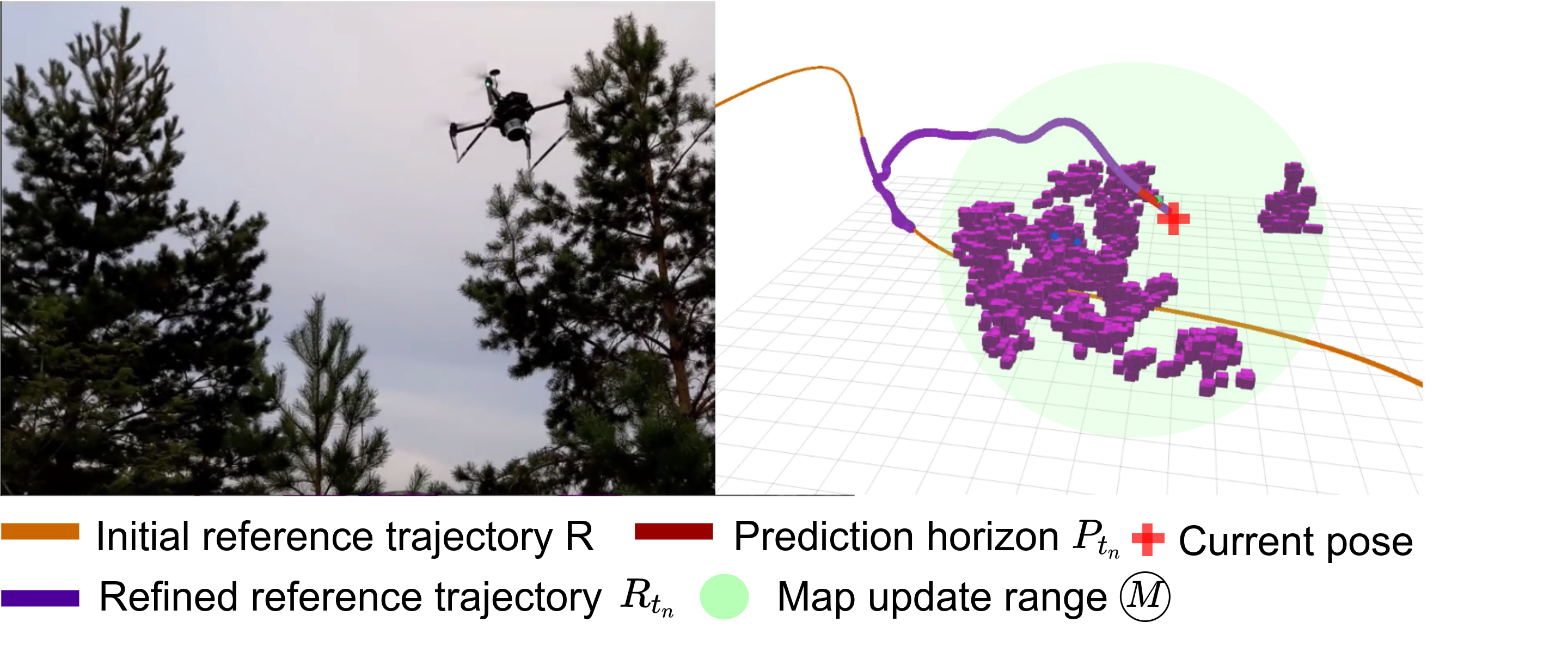}
    \caption{Experiment with the proposed trajectory tracker to showcase how the proposed approach works in real conditions. In this experiment, the total distance of R was about 54m, and the length of $P_{t_n}$ was set to 10}
    \label{fig:testing_method}
    \vspace{-0.2in} 
\end{figure}

This paper proposes an optimization-based approach that solves two trajectories simultaneously, when the first one tries to refine the initial reference trajectory pushing the reference trajectory away from the known obstacles, while the second one generates a near-optimal control policy at every planning step incorporating the refined trajectory (Fig.\ref{fig:testing_method}). Thus, the \textbf{contributions} of this work are as follows:

\begin{enumerate}
    \item Developing a framework for reference trajectory tracking, ensuing safety and dynamic feasibility in which the global planner refines the reference trajectory allowing the local planner to generate a near-optimal control policy quickly at every planning iteration 
    \item Proposing a fast approach, formulated as a convex problem, for pushing the reference trajectory away from obstacle zones, where we implemented a parallel version of Convex Decomposition~\cite{liu2017planning} (Algorithm~\ref{alg:globalPlanner}, line 9) and a simplified approach, as compared to conservative approaches in prior work, for time allocation
    \item Real-world and simulated experiments that showcase the agile flights in various unknown cluttered environments and a new dataset we used for benchmarking our approach with the three other approaches
\end{enumerate}
\section{Related Work}

\begin{figure*}
    \centering
    \vspace{0.2in} 
    \includegraphics[width=0.70\linewidth]{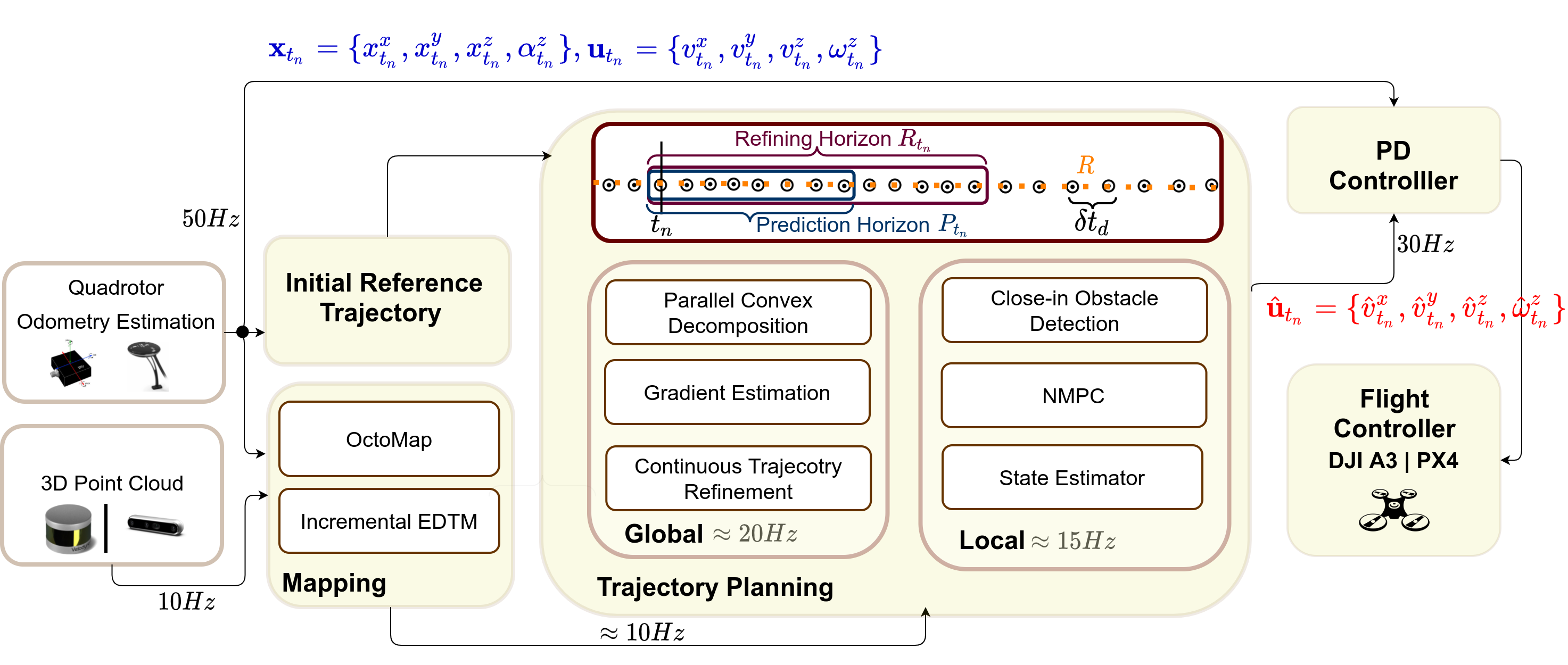}
    \caption{The high-level system architecture of the proposed trajectory tracker. The global and local planners run in parallel as two separate threads while sharing the reference trajectory. Each of the components, i.e., Global, Local, Mapper, average frequencies, are estimated on an NVIDIA Jetson Xavier NX computer (see Fig.~\ref{fig:timestamp}), which is utilized for real-world experiments}
    \label{fig:my_label}
    \vspace{-0.1in} 
\end{figure*}

Most of the recent trajectory tracking approaches are formulated as \textbf{optimization problems} where all the constraints are incorporated in close loop manner. Moreover, such constraints can be embodied as a part of the objective or as a part of the constraints, which can be either soft or hard constraints~\cite{bircher2016receding, oleynikova2016continuous, merkt2019continuous}. Such problem formulation, i.e., employing both objective and constraints, can belong to one of the types: convex~\cite{tordesillas2019faster} or non-convex~\cite{mpc_mc} (non-linear). Non-convex problem formulation is usually computationally expensive. In the recent work of Liu et al.~\cite{liu2018convex}, a successive convex decomposition-based free-space representation, i.e., a series of overlapping polyhedra, was proposed. Such a free-space representation helps to keep  MAV within the free space for the given interval. Specific interval allocation can be calculated in several different ways: let the solver allocate intervals~\cite{landry2016aggressive} or define several intervals prior to solving. Once intervals are allocated, different methods can be used for time allocation (fixed or adaptive)~\cite{liu2017planning}. 

\textbf{Trajectory tracking} problem can be solved during two different planning stages: local or/and global, in which local and global planners can be formulated as two separate or one combined optimization problem. As far as MAVs are concerned, most of the approaches exploit differential flatness property, where the smoothness and dynamic feasibility are estimated by minimizing the L2 norm of velocity difference over the trajectory~\cite{mellinger2011minimum}. However, problem formulation can be complex when enforcing different constraints~\cite{richter2016polynomial}, e.g., obstacle, time, and input constraints. Thus, in the literature, various approaches have been proposed to handle the said constraints. The most primitive paradigm is to use path planning followed by trajectory generation~\cite{kulathunga2020path, kulathunga2020real}, which can be considered as an open-loop problem. However, such approaches fail due to the high computation time, as well as when the environment is highly dynamic and cluttered. To reduce the computation time and have fast reaction time, motion primitive-based local trajectory planning~\cite{mueller2015computationally, zhou2019robust, liu2017search} were proposed. Such approaches are often trapped in local minima. Hence, the objectives of \textbf{global planer and local planner} can differ mainly due to the expected nature (or characteristics) of problem formulation as follows:

The functions and characteristics of a local planner are: it plans for a local near-optimal trajectory based on the currently perceived information within the close vicinity with higher accuracy~\cite{mechali2021observer, tallamraju2019active}; it generates near-optimal control policy in an online fashion in every iteration; it performs trajectory smoothing and feasibility checking to ensure the differential/dynamic constraints; the trajectories are planned consecutively, depending on the way they handle the next set of information that comes in, e.g., how they react to dynamic and static obstacles and how they decide whether to incorporate previous information\cite{baca2018model, 9513301};
long horizon-based trajectory planning may generate wasteful unnecessary long trajectories~\cite{mpc_mc}.

The functions and characteristics of a global planner are: it tries to plan the global near-optimum trajectories~\cite{usenko2017real}; it
constructs the map of the way, either memory-less or fusion-based ~\cite{zhou2019robust}, (a memory-less map does not consider any previous information, but rather considers only current map;  a fusion-based map building does not discard stale data, i.e., previous information, which might be problematic for dynamic obstacles, i.e., some free-known space could be considered as occupied-known space~\cite{kulathunga2020path});
with the known mapping, global planner generates the obstacle-free kinematically feasible trajectory that often ensures the differential/dynamic constraints; it minimizes backtracking and generates efficient trajectories in cluttered environments~\cite{zhou2019robust} in which maintaining a clear picture of the environment is a heavy burden on computational perspective; the computational complexity depends on how much information is incorporated and the way a problem is formulated, e.g., constrained/unconstrained, linear/non-linear optimization problem~\cite{merkt2019continuous, tordesillas2019faster, usenko2017real}.

Therefore, safety, appropriate maneuver, dynamic feasibility, and real-time performance are the main objectives that trajectory tracker must have. The safety and feasible trajectory generation of local or global planners depend on the manner free-known and free-unknown spaces are incorporated. Hence, planning space can be defined to lie within the sensor' FOV or outside of it~\cite{tordesillas2019real, lopez2017aggressive, zhou2020ego}, provided that a series of sensor data has been incorporated for constructing the environment map. When the environment is cluttered, local planners perform poorly due to the uncertainty of instance sensing data. Prediction horizon-based planning is commonly used for such environments, in which local planners can be more conservative compared to global planners. Such a local planner can be formed in multiple ways, e.g., Linear Model Predictive Control (MPC)~\cite{BANGURA201411773}, \textbf{Nonlinear Model Predictive Control (NMPC)}~\cite{mpc_mc} , and Corridor-based Model Predictive  Contouring Control(CMPCC)~\cite{ji2020cmpcc}, based on the necessity and the requirements. The global planner must be less conservative compared to the local planner, e.g., when defined as an unconstrained function minimizer. In the proposed approach, the local planner is formed as an NMPC, whereas the global planner is formed as a box-constrained function minimizer. 

Accurate \textbf{environment mapping} is important to perform robust planning. Out of many, memory-less and fusion-based are the main methods that are used for mapping the environment~\cite{tordesillas2019faster}. Memory-less methods rely on the instantaneous data, i.e., most of the time only on the last sensor reading, whereas fusion-based methods - on the stacking sensor readings in a specific form, e.g., Octomap~\cite{hornung2013octomap}, Voxblox~\cite{oleynikova2017voxblox}, as a map. Such methods may have considerable estimation error and high computation time that depend on the hardware and sensors capabilities. However, estimation error that emerged due to drift and poor sensing measurement can be overcome by resetting the fusion from time to time. Therefore, the latter methods are preferred over the memory-less methods specially for reasoning of cluttered environments due to several reasons, such as limited FOV and the lack of prior information about previous sensing data. Once a map is constructed, Euclidean distance transform mapping (EDTM)~\cite{ragnemalm1993euclidean} can be utilized to estimate the free distance from a considered position. For the mapping, we built instantaneous EDTM on top of Octomap.      

\section{Methodology}

\begin{figure}[ht!]
    \centering
    \vspace{0.1in}
    \includegraphics[width=1\linewidth]{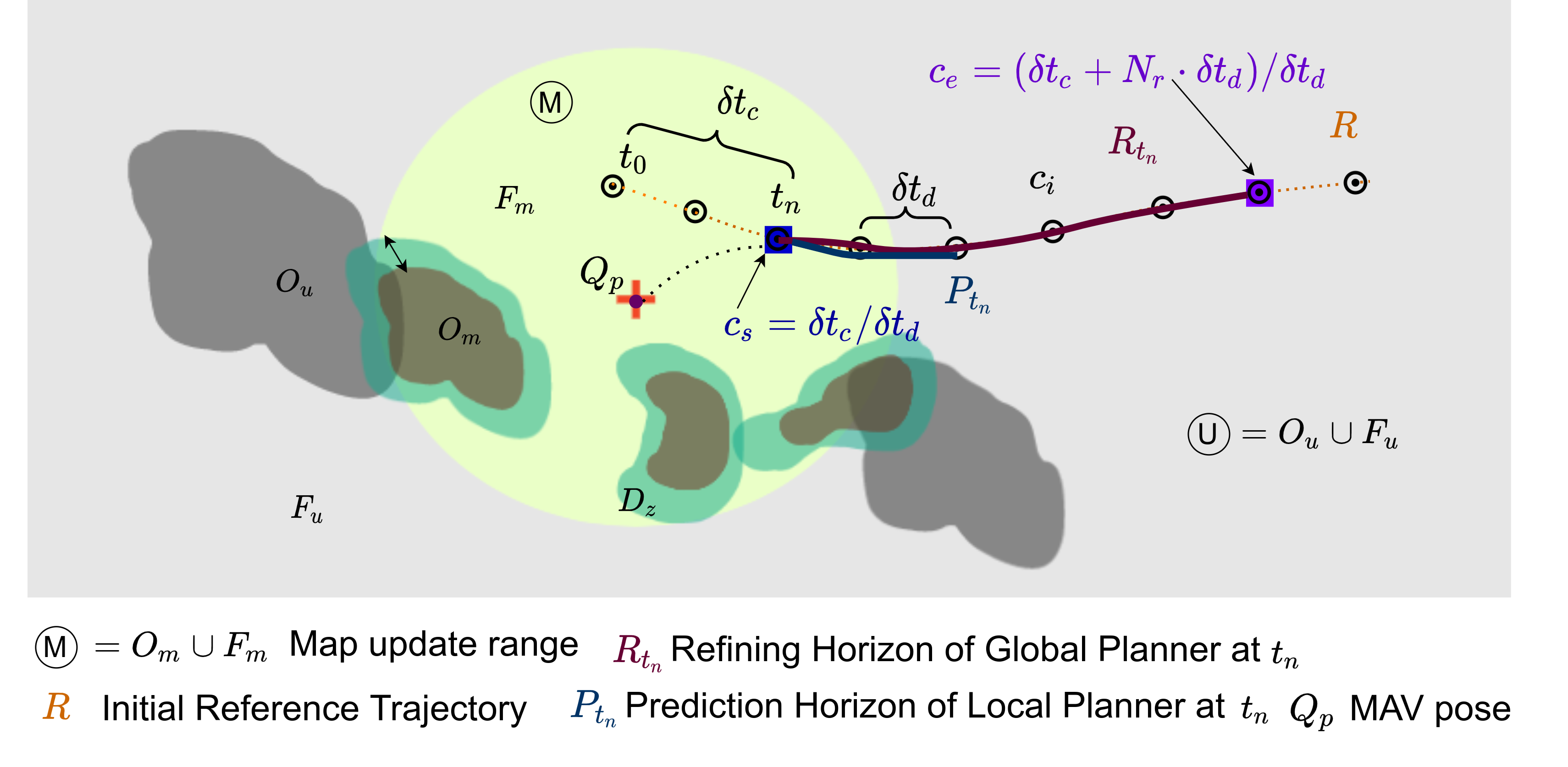}
    \caption{Notion used for defining the reference trajectory tracker and different spaces to the current pose $Q_p$ of MAV}
    \label{fig:problme_definiton}
    \vspace{-0.2in} 
\end{figure}

The proposed approach uses a parallel architecture, which consists of a local and a global planner, and where the global planner pushes the initial reference trajectory away from obstacle zones $O_m$, whereas the local planner generates an optimal control policy for tracking the modified reference trajectory by global planner. The high-level view of the proposed reference trajectory tracker is given in Algorithm \ref{alg:globalPlanner}. A pictorial visualization of the notion that is used throughout the paper is shown in Fig.~\ref{fig:problme_definiton}. The known obstacles and unknown obstacles are defined as $O_m \in \circled{M}$ and $O_u \not\in \circled{M}$, respectively. Similarly, $F_m \in \circled{M}$ and $F_u \not\in \circled{M}$ denote the known free space and unknown free space. Hence, $F_m \cup O_m \subseteq \circled{M}$ and all the unknown region becomes $\mathbb{R}^3 \backslash \circled{M} \subseteq (O_u \cup F_u)$. 
Initial reference trajectory, namely $R$ consists of a set of control points: $c_i, i=0,...,N_c$, where the number of control points is given by $N_c$. For generating $R$, we used the approach proposed in~\cite{mpc_mc}, which is based on uniform bspline. The initialization time of the trajectory planning and current time, where the desired pose on the reference trajectory should lie, are denoted by $t_0$ and $t_n$, respectively. For a given time $t_n$, starting and finishing control points are retrieved with respect to $ c_s \subseteq [0, N_c)$ and $c_e \subseteq (c_s, N_c-1]$ indices. The time difference between two consecutive control points, i.e., $c_i$ and $c_{i+1}$, is defined as $\delta_{t_d}$, which was set to 0.05s in our experimental setup. The actual discretization time interval of continuous system dynamics $\delta_{t_c}$ was set to 0.05. Besides, $\delta_{t_c}$ and $\delta_{t_d}$, which both can be the same or slightly different from each other, can be configured. The number of control points within the $R_{t_n}$, namely, refining horizon, is denoted by $N_r$. The avoidance distance, which is the minimum free distance $D_z$ allowed in between MAV and the closest obstacle to MAV.

\begin{algorithm}
\caption{Reference trajectory tracker}\label{alg:globalPlanner}
\begin{algorithmic}[]
\State $\mathbf{Inputs}$: at time $t_n$, $\textcolor{reds}{R_{t_n}}$: reference trajectory to be refined, $Q_p$: current pose of MAV, $\textcolor{blue}{P_{t_n}}$: trajectory to be tracked , $M_{t_n}$: EDT map of the environment 
\Statex $\mathbf{Outputs}$: $\textcolor{reds}{R_{t_n}}$: refined reference trajectory; $\textcolor{blue}{P_{t_n}} \in \textcolor{reds}{R_{t_n}}$, $v_x, v_y, v_z, \omega_z$ : control command to maneuver  MAV; 
 \Statex \hrulefill
\Procedure{Global Planner}{}
\State $\textcolor{reds}{R_{t_n}} \gets <Q_p, \textcolor{reds}{R_{t_n}}>$
\State $S_o \gets$ CheckingOccupiedSegments($\textcolor{reds}{R_{t_n}}, M_{t_n}$)
\If{$S_o >0$}
    \For{$i \gets S_o$}
        \State $A_i, b_i \gets $ ParallelConvexDecomposition($S^i_o$) 
        \State $S_*^i \gets $ FindPushingDirections($S^i_o, A_i, b_i$)
        \State $\textcolor{reds}{R_{t_n}} \gets$ CalculateGradients($S_*^i, \textcolor{reds}{R_{t_n}}$)
    \EndFor
\EndIf
\State return $\textcolor{reds}{R_{t_n}} \gets$ ApplyBoxConstraintOptimization($\textcolor{reds}{R_{t_n}}$)
\EndProcedure
\Statex \hrulefill
\Procedure{Local Planner}{}
\State $\textcolor{blue}{P_{t_n}} \gets <Q_p, \textcolor{blue}{P_{t_n}}>$
\State $C_o \gets$ GetCloseInObstacles($\textcolor{blue}{P_{t_n}}, M_{t_n}$)
\State return $<v_x, v_y, v_z, \omega_z> \gets$ ApplyNMPC($\textcolor{blue}{P_{t_n}}, C_o$)
\EndProcedure
\end{algorithmic} 
\end{algorithm}

\subsection{Global Planner}
The proposed approach consists of two planning stages: local and global. The local planner is designed as a constraint nonlinear optimization problem (NLP), specifically an NMPC. The computation time of NMPC increases when the number of constraints increases, i.e., reference trajectory lies closer or within the obstacles. Such behaviour can lead to local minima and cannot find near-optimal control policy to avoid close-in obstacles. Hence, the global planner does refine the initial reference trajectory in parallel with the local planner to push the reference trajectory away from the obstacle zones. 
Hence, the proposed global planner is formulated as follows:
\begin{equation}\label{alg:global_opt}
    \begin{aligned}
          J = \lambda_{smooth}J_{smooth} + \lambda_{obs}J_{obs} + \lambda_{feasibility}J_{feasibility}, \\
    \end{aligned}
\end{equation} where $\lambda_{*}, *\in \{ smooth, obs, feasibility\}$ are weight parameters were set as 0.2, 0.6, and 0.2, respectively. $ \lambda_{smooth}$ and $ \lambda_{feasibility}$ were set to low values mainly due to relax smoothness and feasibility adjustment compared to obstacle avoidance adjustment by putting high penalty weight $\lambda_{obs}$. In the following sub-sections, formulation of global planner components is explained adhering to Algorithm.~\ref{alg:globalPlanner}.

\subsubsection{Finding Pushing Direction}\label{sec:pusing_points}
Some of the control points in $R_{t_n}$ can occur within the obstacles zones. Hence, control points that lie within the obstacle zone $O_m$ must be pushed towards an obstacle-free zone $F_m$. Let $S_o^i \in R_{t_n}$ be the $i^{th}$ segment within $R_{t_n}$ to be modified. $S_o^i$ consists of a set of control points $\mathbf{c}_j \in S_o^i, j=0,...,N^{seg}_i$, where $N^{seg}_i$ is the number of control points in $S_o^i$. Thus, pushing direction of each control point is determined by solving the following convex problem for each segment:
\begin{equation}\label{alg:pusing_direction}
    \begin{aligned}
        \min_{\mathbf{p}_0,..., \mathbf{p}_{n}} & \lambda_1 t_1 + \lambda_2 t_2+ \lambda_3 t_3 \\
        \textrm{s.t.} & \quad   A\mathbf{p}_j \leq b, \\
        \quad &  \left \| \mathbf{p}_0 - \mathbf{c}_{0} \right \|_2 \leq t_1 ,\\ 
         \quad &  \left \| \mathbf{p}_{n} - \mathbf{c}_{n} \right \|_2 \leq t_2, \\ 
          \quad &   \sum _{k=1}^{n-1}\left \| \mathbf{p}_{k+1} - \mathbf{p}_{k} \right \|_2 \leq t_3 ,\\ 
    \end{aligned}
\end{equation} where $A$ and $b$ represents the free space $F_m$ as a convex polyhedron from $\mathbf{c}_0$ to $\mathbf{c}_{n}$ in $S_o^i$ (Sec.\ref{sec:paralle_decompose}). For the $i^{th}$ segment, $n = N^{seg}_i$. The regularization parameters: $\lambda_1 = 0.8, \lambda_2 = 0.8$, and $\lambda_3 =  0.6$, were set in a way to provide more bias on start and end control points compared to middle control points.  $\mathbf{p}_0,..., \mathbf{p}_{N^{seg}_i}$ construct the updated segment $S_*^i$ corresponding to $S_o^i$ that will be used for finding each control point's gradient direction (Sec.~\ref{sec:calgradients}). 

\subsubsection{Parallel Convex Decomposition}\label{sec:paralle_decompose}
To further reduce the computation time, we have implemented the parallel version of Convex Decomposition~\cite{liu2017planning} (Algorithm~\ref{alg:globalPlanner}, line 9). Once desired control points are identified, i.e., $\mathbf{c}_j \in S_o^i, j=0,...,N^{seg}_i$ (Sec.\ref{sec:pusing_points}), check for intermediate control points that are in $F_m$. Convex decomposition is applied to such successive control points in parallel that result in the free space in the form of H-rep  $Ax \leq b$ for each $S_o^i$. 

\subsubsection{Calculating Gradients}\label{sec:calgradients}
\begin{figure}[ht!]
    \centering
     \vspace{0.05in} 
    \includegraphics[width=0.9\linewidth]{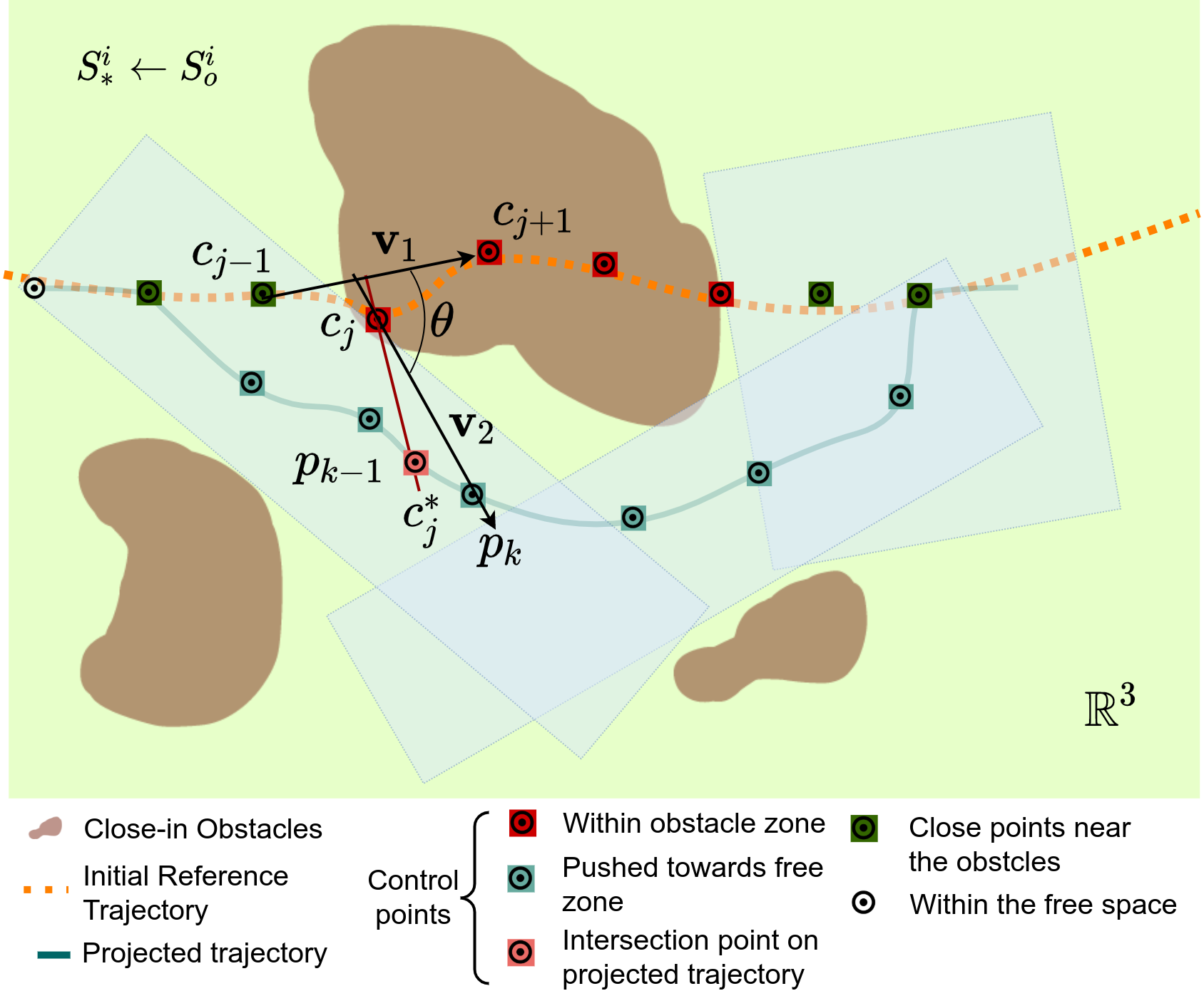}
    \caption{Pushing control points that are within the obstacle zone, towards the free space. Projected control points segment ($S_*^i$) is obtained as explained in Sec.~\ref{sec:pusing_points} utilizing $S_o^i$ for $i^{th}$ segment. $c_j^*$ depicts gradient vector corresponding to $c_j$}
    \label{fig:gradient_estimation}
     \vspace{-0.08in} 
\end{figure}

The objective of $J_{obs}$ is to push each $S_o^i, i=0,...,N^{seg}$ segment towards the obstacle-free zone. $N^{seg}$ is the number of segments that are within the obstacle zone for the considered refine trajectory segment $R_{t_n}$, at time $t_n$. For the $i^{th}$ segment, by knowing $S_o^i$, $S_*^i$ can be determined (\ref{sec:pusing_points}). To find each gradient direction vector that crosses the $  \mathbf{c}_j \perp S_*^i,\; j=0,..,N^{seg}_i$, let $\mathbf{v}_1 = \mathbf{c}_{j+1} - \mathbf{c}_{j-1}$ be the approximated direction vector along $\mathbf{c}_j$, and  $\mathbf{p}_k \in S_*^i$ be the control point that intersects $\mathbf{v}_1$ ( Fig.~\ref{fig:gradient_estimation}). Then, the corresponding direction vector $\mathbf{v}_2$ can be defined as $\mathbf{p}_{k} - \mathbf{c}_j$. By calculating angle $\theta = cos^{-1}(\mathbf{v}_1 \cdot \mathbf{v}_2/ \left \| \mathbf{v}_1 \right \|_2 \left \| \mathbf{v}_2 \right \|_2)$ between $\mathbf{v}_1$ and $\mathbf{v}_2$, the optimal value of k can be determined as provided in Algorithm ~\ref{alg:gradient_vector}. Thus, the gradient vector that corresponds to $c_j$ can be fully determined as:
\begin{equation}
\begin{split}
   &  \mathbf{c}_j^{grad} = (\mathbf{c}_j^*-\mathbf{c}_j)/\left \|  \mathbf{c}_j^* -\mathbf{c}_j\right \|_2, \\
   & \mathbf{c}^*_j = \mathbf{p}_k + \frac{(\mathbf{p}_{k} - \mathbf{p}_{k-1})(\mathbf{v}_1 \cdot (\mathbf{c}_j-\mathbf{p}_k))}{\mathbf{v}_1\cdot (\mathbf{p}_{k}-\mathbf{p}_{k-1})}
\end{split}.
\end{equation}


\begin{algorithm}
\caption{Estimation of direction vectors pushing the control points towards the free space.}\label{alg:gradient_vector}
\begin{algorithmic}[0]
\Statex $\mathbf{Inputs}$: at time $t_n$, $\textcolor{reds}{R_{t_n}}$: reference trajectory to be refined, $N^{seg}$: segments indices to be refined within $\textcolor{reds}{R_{t_n}}$
\Statex $\mathbf{Outputs}$: $\textcolor{reds}{R_{t_n}}$: after adding, gradient vector corresponds to each control point
\Statex \hrulefill
\Procedure{GradientEstimation}{}
\For{$i \gets 0 \quad to \quad N^{seg}$}
    \For{$j \gets 1 \quad to \quad N_{i}^{seg}$}
        \State $k \gets N^{seg}_i/2$
        \State $\mathbf{v}_1 = \mathbf{c}_{j+1} - \mathbf{c}_{j-1}, \; \mathbf{v}_2 = \mathbf{p}_{k} - \mathbf{c}_j$
        \State $  \mathbf{c}_j \in S_o^i,\; \mathbf{p}_k \in S_*^i$
        \State $val = previous\_val \gets \mathbf{v}_1 \cdot \mathbf{v}_2$
        \While{$k \geq 0 \quad and \quad k < N_i^{seg}$}
            \State $k \gets \left\{\begin{matrix}  k--, \quad if \: val \leq 0 \\  k++, \quad otherwise \end{matrix}\right.$
            \State $val \gets \mathbf{v_1}\cdot \mathbf{v_2}$
            \If{$val \cdot previous\_val \leq 0 $}
               \State  $\mathbf{c}^*_j \gets \mathbf{p}_k + \frac{(\mathbf{p}_{k} - \mathbf{p}_{k-1})(\mathbf{v}_1 \cdot (\mathbf{c}_j-\mathbf{p}_k))}{\mathbf{v}_1\cdot (\mathbf{p}_{k}-\mathbf{p}_{k-1})}$
               \State $\delta d \gets \left \|  \mathbf{c}_j^* -\mathbf{c}_j\right \|_2$
               \State $ \mathbf{c}_j^{grad} = \frac{\mathbf{c}_j^*-\mathbf{c}_j}{\left \|  \mathbf{c}_j^* -\mathbf{c}_j\right \|_2}$
            \EndIf
        \EndWhile
    \EndFor 
\EndFor
\State return $\textcolor{reds}{R_{t_n}}$
\EndProcedure
\end{algorithmic} 
\end{algorithm} 

Once gradient vectors are estimated, $J_{{obs}}$ is determined (\ref{eq:obs_cost}), which is defined as a continuously differentiable exact penalty function.

\begin{equation}\label{eq:obs_cost}
    \begin{split}
        & J_{obs} = \Sigma_{i=d}^{N_r-d} J_{{obs}_i},\\
        & J_{{obs}_i} = \mathbf{v}_i \cdot dis_{e}^3, \quad \frac{\partial J_{{obs}_i}}{\partial \mathbf{c}_i} = -3 \cdot dis_{e}^2 \cdot \mathbf{c}_i^{grad}, \\
    \end{split}
\end{equation} where $dis_{e} = D_z - (\mathbf{c}_i-\mathbf{c}_i^*) \cdot \mathbf{c}_i^{grad}$ and $ \mathbf{v}_i = \mathbf{c}_{i+1} - \mathbf{c}_i$, and avoidance distance $D_z$ was set to 0.8m (distance must be higher than the radius of the MAV) in our study.

\subsubsection{Dead Zone Recovery}
The map construction is not precise when the depth sensor has a small FoV. Moreover, EDTM building takes a considerable amount of time when the environment is cluttered. Therefore, the local planner may generate control commands that lead to quadrotor $Q_p$ maneuvers into the $D_z$ (Fig.~\ref{fig:problme_definiton}) zone. In such situations,
free space segmentation (Sec.\ref{sec:paralle_decompose}) does not provide correct constraints set $A,b$. Hence, the proposed approach (\ref{alg:pusing_direction}) fails to estimate control points  $\mathbf{p}_0,...,\mathbf{p}_{N_i^{seg}}$ appropriately, i.e., estimated control points may lie in the extreme ends ($\mathbf{c}_0, \mathbf{c}_{N_{i}^{sec}}$) of the provided trajectory $S_o^i$, provided that $R_{t_n}$ is not dynamically feasible. Hence, the objective is to consider whole $R_{t_n}$ rather than each segment separately, followed by the free space segmentation. Thus, the following recovery mechanism is proposed to push the $R_{t_n}$ away from $O_m$. The recovery mechanism is executed only when the (\ref{alg:pusing_direction}) is failed. 

\begin{equation}\label{eq:deadzone}
    \begin{aligned}
     \min_{\mathbf{p}_1,..., \mathbf{p}_{N_r}} & \sum_{l=1}^{N_r} q_l\\
        \textrm{s.t.} & \quad   A_r(\mathbf{p}_l+\mathbf{c}_l) \leq b_r, \;  \left \| \mathbf{p}_l \right \|_2 \leq q_l, \; l=1,...,N_r,\\
    \end{aligned}
\end{equation} where $c_l, l=1,...,N_r$ are the control points to be pushed. The recovered control points are determined by $\mathbf{p}_l+ \mathbf{c}_l$, $N_r$ is the number of control points at time $t_{n}$ in $R_{t_n}$, and $A_r$ and $b_r$ are obtained by giving $R_{t_n}$ to Algorithm~\ref{alg:globalPlanner}, line 9.     

\subsubsection{Smoothing}
We have employed a velocity controller since the proposed trajectory tracker targets low-speed maneuvers. Hence, higher-order components, i.e, acceleration, jerk, snap, should be minimized, which causes effects such as vibrations. However, we decided to minimize only acceleration components without considering higher-order components, e.g., jerk, snap. We have formulated $J_{smooth}$ minimizing both acceleration and jerk components as well as only considering acceleration components. However, adding jerk did not affect $J_{smooth}$ considerably. Thus, $J_{smooth}$ was formulated only as minimizing the acceleration components:
\begin{equation}
    \begin{split}
        & J_{{smooth}_i} = \mathbf{a}_i^{\top}\mathbf{a}_i, \; \; \; \frac{\partial J_{{smooth}_i}}{\partial \mathbf{c}_i} =  2 \frac{\partial \mathbf{a}_i}{\partial \mathbf{c}_i}, \\
    \end{split}
\end{equation} where $ \partial \mathbf{a}_i/\partial \mathbf{c}_i = 1, \; \partial \mathbf{a}_i/\partial \mathbf{c}_{i+1} = -2, \; \partial \mathbf{a}_i/\partial \mathbf{c}_{i+2} = 1 $. $\mathbf{a}_i, \mathbf{v}_i, \mathbf{c}_i \in \mathbb{R}^3$ are respectively acceleration ($\mathbf{a}_i = \mathbf{c}_{i+2} - 2\mathbf{c}_{i+1} + \mathbf{c}_i$), velocity ($\mathbf{v}_i = \mathbf{c}_{i+1} - \mathbf{c}_i)$, and control point at $i^{th}$ index of $R_{t_n}$.

\subsubsection{Feasibility}
To ensure the refined trajectory, namely, $R_{t_n}$, which is dynamically feasible for the maneuver, objective function penalizes the velocity and acceleration components only when their limits exceed the min and max, as follows: 
    \begin{equation}
       \begin{split}
         & J_{{feasibility}_i} =  (\mathbf{v}_i\oplus \mathbf{v}_{max})^{\top}(\mathbf{v}_i\oplus \mathbf{v}_{max})\cdot \frac{1}{\delta^2} \\
 & \quad \quad \quad \quad \quad \quad + (\mathbf{a}_i \oplus \mathbf{a}_{max})^{\top}(\mathbf{a}_i \oplus \mathbf{a}_{max}) 
       \end{split}
   \end{equation} where the operator $\oplus$ is defined as 
   \begin{equation}
        \oplus	 = \left\{\begin{matrix} - & if \; \mathbf{v}_i > \mathbf{v}_{max} \; || \; \mathbf{a}_i > \mathbf{a}_{max}  \\ + & if \; \mathbf{v}_i < -\mathbf{v}_{max} \; || \; \mathbf{a}_i < -\mathbf{a}_{max} \\ not \; considering &    otherwise \end{matrix}\right.,
   \end{equation} where allowed maximum velocity and acceleration components are given by $\mathbf{v}_{max} \in \mathbb{R}^3$ and $\mathbf{a}_{max} \in \mathbb{R}^3$, respectively. When the velocity and acceleration components are within the allowed range, there will be no added cost. Once objective function $J$~(\ref{alg:global_opt}) was formed, we have used L-BFGS-B (Limited-memory Broyden Fletcher Goldfarb Shanno Box-constrained algorithm)~\cite{LBFG} for solving J. Subsequently, Mosek solver~\cite{Mosek} was employed to solve~(\ref{alg:pusing_direction}) and (\ref{eq:deadzone}).

\subsection{Local Planner}
At time $t_n$, $ P_{t_n} = [\mathbf{Q}_p, \mathbf{c}_{t_n}, \mathbf{c}_{t_n+1}, ..., \mathbf{c}_{t_n+N_p}] \subseteq R_{t_n}$ forms the reference trajectory for the given prediction horizon, $N_p$.  The local planner generates the optimal control to maneuver the quadrotor considering close-in obstacles $g_2(\mathbf{w})$ and system dynamics $g_1(\mathbf{w})$ where $\mathbf{w} = [\textbf{u}_{t_n}, \hdots , \textbf{u}_{t_n+N_p-1}, \textbf{x}_{t_n}, \hdots, \textbf{x}_{t_n+N_p}]$. Hence, the objective of local planner is to optimize both control inputs and states simultaneously. Such an objective can be designed using multiple shooting technique as follows:  

\begin{equation}\label{eq:nmpc}
\begin{aligned}
J_{P} (\textbf{x}, \textbf{u})_{t_n} & = \sum_{l=0}^{N_p}{\left \| \textbf{x}_{t_n+l} -\textbf{c}_{t_n+l} \right \|_Q^2 + \left \| \textbf{u}_{t_n+l} -\textbf{v}^{ref}_{t_n+l} \right \|_R^2} \\
 \min_{\mathbf{w}} \quad & J_{P} (\textbf{x}, \textbf{u})_{t_n} \\
\textrm{s.t.} \quad  & g_1(\mathbf{w}) = 0 , \quad g_2(\mathbf{w}) \leq 0 \\
  \quad & \mathbf{x}_{min} \leq \textbf{x}_{t_n+l} \leq \mathbf{x}_{max} \quad \forall 0 \leq l \leq N_p \\
  \quad & -\mathbf{v}_{max} \leq \textbf{u}_{t_n+l} \leq \mathbf{v}_{max} \quad \forall  0 \leq l \leq N_p-1 ,
\end{aligned}
\end{equation} At every planning cycle, local planner gets $Q_p, \mathbf{x}_{t_n}$, and $\mathbf{u}_{t_n}$ as an input and estimates the optimal control policy, i.e., $\mathbf{\hat{u}}_{t_n} = \{ \hat{v}_{t_n}^x, \hat{v}_{t_n}^y, \hat{v}_{t_n}^z, \hat{\omega}_{t_n}^z\}$, where $\hat{v}_{t_n}^\mu, \mu \in x, y, z$ denotes velocity on each $\mu$ direction, and yaw angle around z axis is given by $\hat{\omega}_{t_n}^z$. The local planner is adopted from our previous work, where the explanation of $g1$, $g2$ and  $\mathbf{\hat{u}}_{t_n}$ is detailed~\cite{mpc_mc}. 

To conclude, as summarized in Algorithm.~\ref{alg:globalPlanner}, this section explained how the proposed trajectory tracker is formulated. In the following section, the qualitative and quantitative analysis of the proposed approach is provided.  
\section{EXPERIMENTAL PROCEDURE AND RESULTS}

\begin{figure}[ht]
    \centering
     \vspace{0.2in} 
\includegraphics[width=0.5\linewidth]{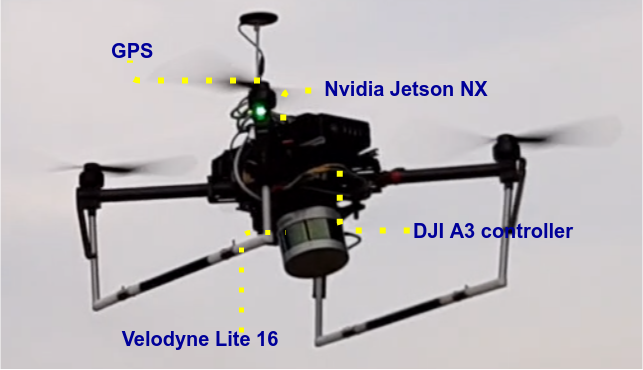}
    \caption{The experimental prototype MAV (DJI M100) was used in real-world experiments. The proposed approach runs on the on-board computer (Nvidia Jetson NX) and sends control commands to A3 controller}
    \label{fig:hardware_setup}
    \vspace{-0.1in} 
\end{figure}

The experiment prototype of MAV DJI M100 (Fig.~\ref{fig:hardware_setup}) is equipped with the following components: Velodyne Lite 16 lidar for reasoning the MAV's surrounding environment; Nvidia Jetson NX computer for online computations. Offline computations and the simulated experiments were carried out on an Intel i9-9900K (16) @ 5 GHz computer. Hence, the timing breakdowns were measured for both real-world and simulated experiments by those two computers. The simulated experiments were performed in a Gazebo environment. For the simulated and real-world experiments, PX4~\cite{meier2015px4} and DJI A3 controllers were employed, respectively. 

\begin{figure}[ht!]
    \centering
    \includegraphics[width=0.8\linewidth]{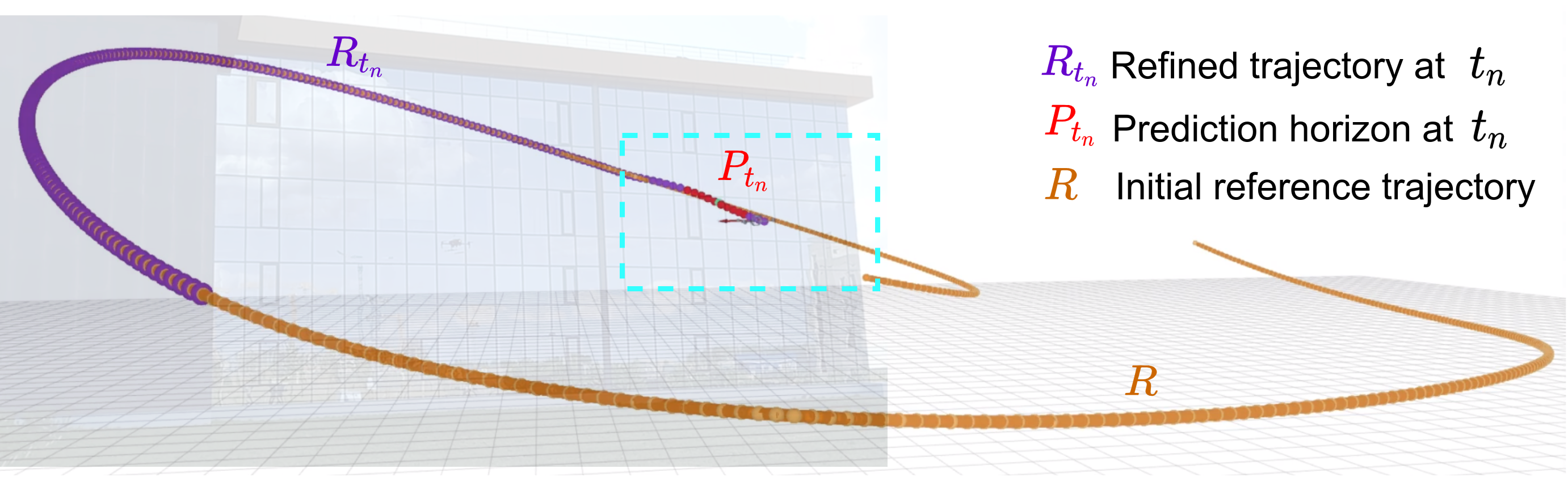}
    \caption{Experimental results for real-world tracking accuracy without considering obstacles. Estimated tracking error is less than 1m during the whole experiment}
    \label{fig:trajectrory_error}
    \vspace{-0.1in} 
\end{figure}

The first experiment\footnote[2]{\label{freespace} tracking accuracy without considering obstacles:~\url{https://www.youtube.com/watch?v=pKVeGdr8crU}} was aimed to estimate reference trajectory tracking error without refining. Such trajectory tracking can be directly used in, for example, cinematography. Moreover, this is a way to check the local planner~(\ref{eq:nmpc}) is able to track the reference trajectory that the proposed global planner provides. As shown in Fig.~\ref{fig:trajectrory_error}, we estimated position estimation error $\left | p - p^{ref} \right |_{2}$ between the tracked trajectory $p$ and the reference trajectory $p^{ref}$. The fusion of GPS position, IMU data, and vehicle velocity is used to estimate $p$, whereas $p^{ref}$ is the output of local planner. The mean estimation error $\overline{\left | p - p^{ref} \right |}_{2}$ was less than 1m during the whole flight in which velocity varied in between -1.2 m/s to 1.2 m/s.  

\begin{figure}[ht!]
    \centering
    \vspace{0.2in} 
    \includegraphics[width=0.9\linewidth]{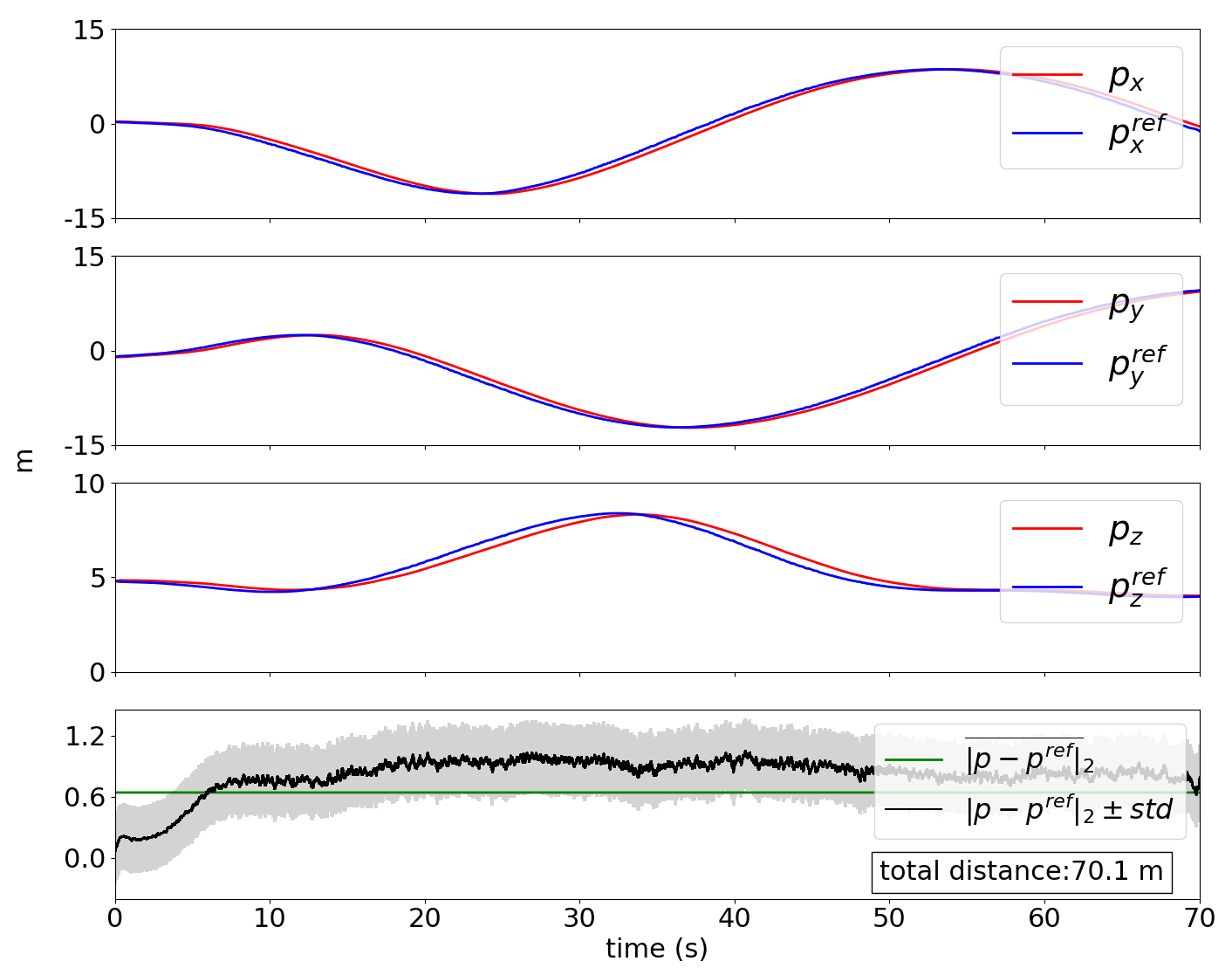}
    \caption{The trajectory tracking error for real-world experimental results on tracking the reference trajectory shown in Fig.~\ref{fig:trajectrory_error}}
    \label{fig:reference_error}
    \vspace{-0.1in} 
\end{figure}

\begin{table*}[h]
\vspace{0.2in} 
\caption{Comparative analysis of the proposed approach and the three other approaches for checking goal-reaching accuracy. All the metrics were obtained during experiments on 12 different environments while keeping the same start and goal poses}
\label{t:comparision}
\begin{tabular}{|l|l|l|l|l|l|}
\hline
\multirow{2}{*}{Algorithm} & \multirow{2}{*}{Success Fraction (SF)} & \multirow{2}{*}{Mean Computation Time (MCT) in seconds } & \multicolumn{3}{l|}{Distance Estimation (m)} \\ \cline{4-6} 
                           &                                                                                   &                                                                                         & Mean         & Max         & Min         \\ \hline
\cite{kulathunga2020real} RRT* max\_allowed\_iterations=10000          & 0.41      &   2.4 &    112.56   &  198.45  &   91.04\\ \hline          \cite{mpc_mc} $P_{t_n}=20, update\_range= 5m$                        & 0.58 & 0.397 & 93.5  &   154.67  &  68.45 \\ \hline                      
\cite{mpc_mc} $P_{t_n}=40, update\_range= 5m$                        & 0.66  & 0.441 &  98.91  &  201.56   & 78.23 \\ \hline
\cite{tordesillas2019faster} $N_{whole} = N_{safe}=6$  $max\_poly = 3$ & 0.83  & $\approx 0.01$ &  54.56  &  81.45   & 49.39 \\ \hline
\cite{tordesillas2019faster} $N_{whole} = N_{safe}=12$ $max\_poly = 6$& \textbf{0.83}   & $ \approx \textbf{ 0.01} $  &  53.78  &  78.67   & 47.45 \\ \hline
Proposed $P_{t_n}=10, update\_range= 4m$ & 0.91    & $0.04 \pm 0.01$  &  56.78  &  68.78   & 49.86 \\ \hline
Proposed $P_{t_n}=15, update\_range= 6m$ & \textbf{1.0}   & $\mathbf{0.03}\pm \textbf{0.01}$ &   54.89  &  66.80   & 48.67 \\ \hline
\end{tabular}
\begin{tablenotes}
     \item[1]   $N_{whole}, N_{safe}$ : the number of discretization points in the whole and safe trajectory,  $max\_poly$: maximum number of polydrons to represent the free space, $P_{t_n}$: NMPC prediction horizon length 
\end{tablenotes}
\end{table*}

\begin{figure}[ht!]
    \centering
    \includegraphics[width=1\linewidth]{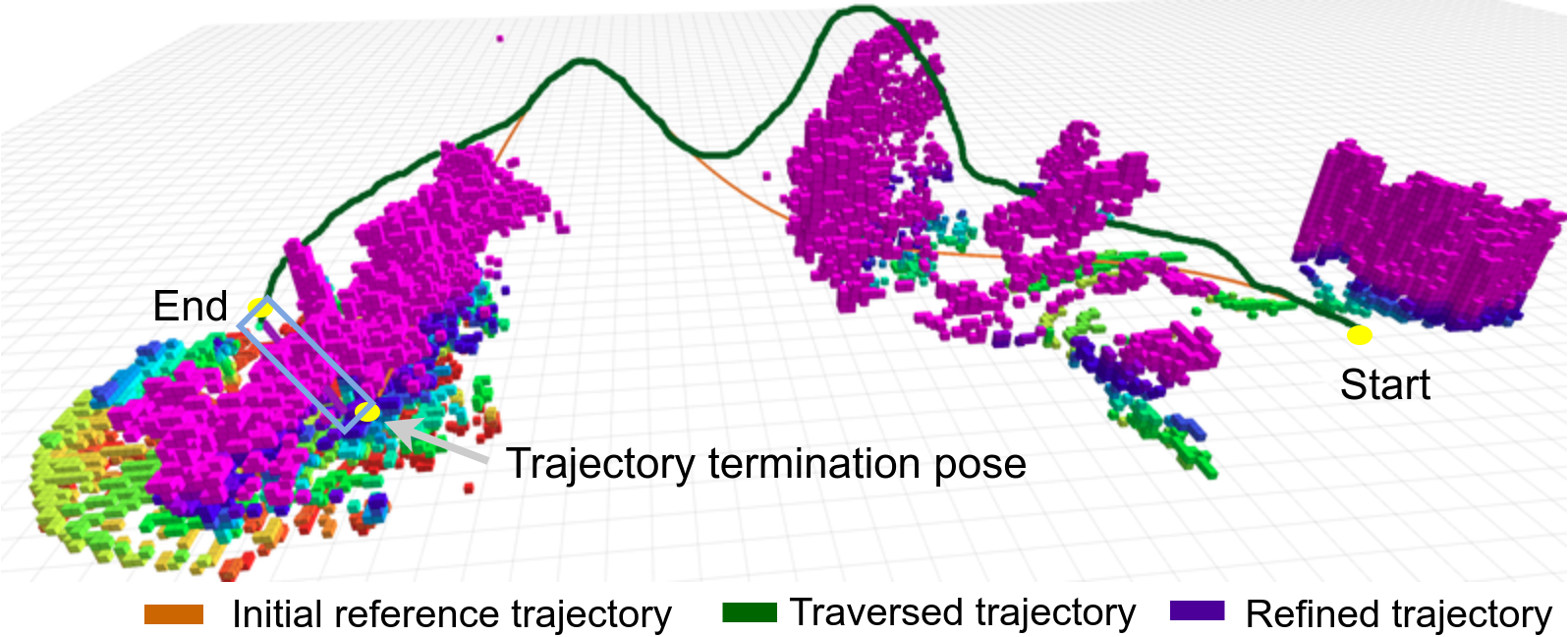}
    \caption{Showcasing the behaviour of the proposed trajectory tracker in a challenging environment}
    \label{fig:test_1_map}
\end{figure}

The second experiment was devoted to demonstrating the behaviour of the proposed approach in a real-world condition\footnote[3]{\label{realtest} behaviour of the proposed approach in a challenging environment~\url{https://youtu.be/g6xHvkcrYcQ}}, where the initial reference trajectory passes through a cluttered environment followed by open space and back to a cluttered environment where the terminal pose was placed in an obstacle zone. Map update range ($update\_range$) was kept 4m from the center of the MAV and max speed set to 0.6m/s for the safety of MAV. Total flight time was around 150s and the distances of initial reference trajectory and traversed trajectory were 54.3m and 79.8m, respectively. Since trajectory termination pose was within the obstacles, trajectory tracker terminates early(see Fig.~\ref{fig:test_1_map}). Such a behaviour is due the fact that the global planner was designed as a box-constraint function minimizor, whereas local planner was designed as a constraint NLP. Hence, the local planner terminated correctly, though the global planner completely failed to refine, which is in fact true. 

\begin{figure}[ht!]
    \centering
    \includegraphics[width=0.9\linewidth]{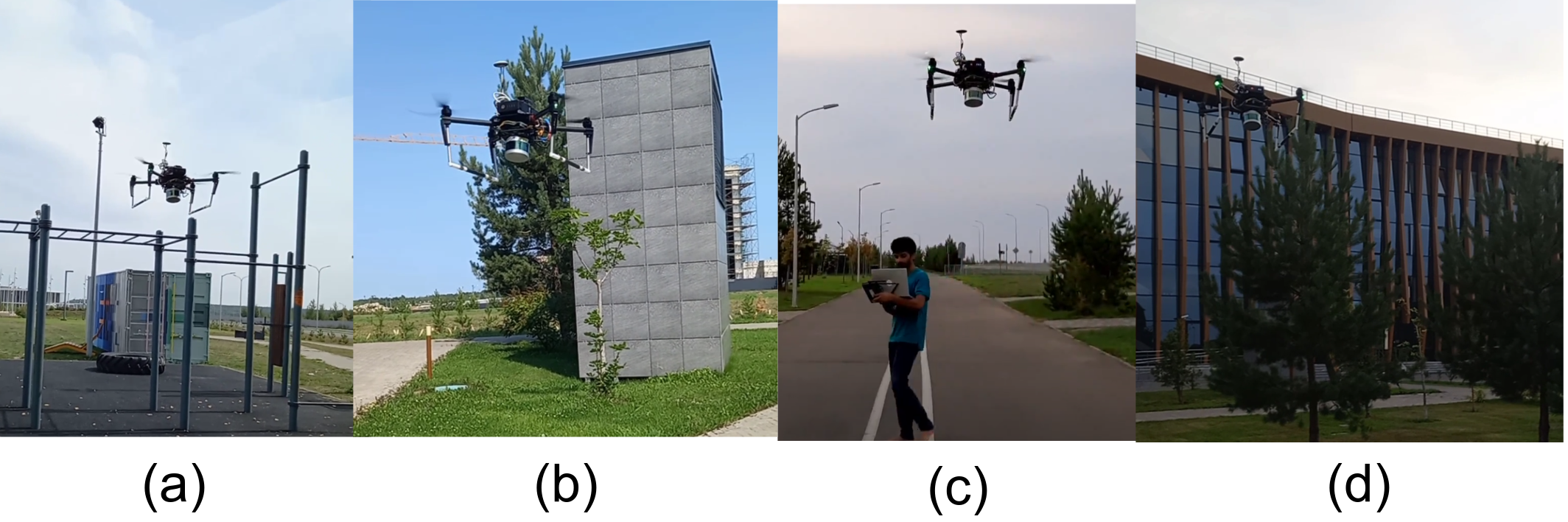}
    \caption{ Different scenarios for real-world experiments: static (a,b,d) and dynamic (c) were used to estimate the run-time breakdowns of the proposed trajectory tracker (Fig.\ref{fig:timestamp})}
    \label{fig:run_time_estimation}
     
\end{figure}

\begin{figure}[ht!]
    \centering
    \includegraphics[width=1\linewidth]{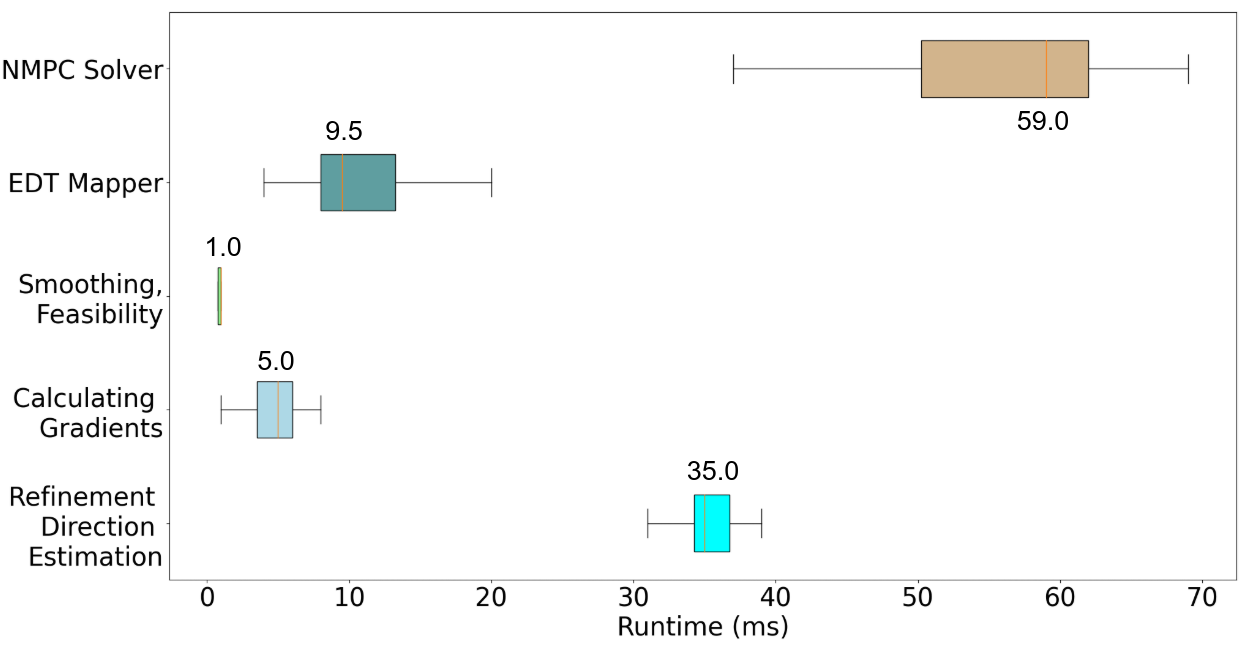}
    \caption{Estimation of mean computation time (run-time) of the proposed approach, i.e., time break down of each sub components, in real-world scenarios (see Fig.\ref{fig:run_time_estimation}) }
    \label{fig:timestamp}
    \vspace{-0.18in}
\end{figure}

In the third experiment\footnote[4]{\label{realtest} experiments used for estimating the run-time~\url{https://www.youtube.com/watch?v=jyDe5BSigm8}}, we conducted four different real-world tests to estimate the run-time breakdown (mean computation time) in the average case. Three out of four tests were performed in static environments: open area with small obstacles, open area with sizeable obstacles, and a cluttered environment) and the fourth experiment was performed in a dynamic environment ( Fig.\ref{fig:run_time_estimation}). Reference trajectories of each of them were completely different from each other. However, we fixed the trajectory tracking duration to 90s. Afterwards, run-time breakdown (Fig.\ref{fig:timestamp}) was estimated based on three sub-modules: NMPC solver (main force of the local planner), EDT mapper (utilize both local and global planner), and main parts of the global planner (smoothing, feasibility, calculating gradients, and finding pushing directions).  The objective was to understand how the run-time of each of the listed sub-modules is affected due to environmental changes. Since those three modules were executed in parallel, local and global planners have mean computation times of approximately 0.06s (15Hz) and 0.05s (20Hz), respectively.

\begin{table*}[H]
\vspace{0.2in} 
\caption{Comparative analysis of the proposed approach and the three other approaches for checking goal-reaching accuracy.  All the metrics have been obtained from experiments on 12 different environments while keeping the same start and goal poses}
\label{t:comparision}
\begin{tabular}{|l|l|l|l|l|l|}
\hline
\multirow{2}{*}{Algorithm} & \multirow{2}{*}{Success Fraction (SF)} & \multirow{2}{*}{Mean Computation Time (MCT) in seconds } & \multicolumn{3}{l|}{Distance Estimation (m)} \\ \cline{4-6} 
                           &                                                                                   &                                                                                         & Mean         & Max         & Min         \\ \hline
\cite{kulathunga2020real} RRT* max\_allowed\_iterations=10000          & 0.41      &   2.4 &    112.56   &  198.45  &   91.04\\ \hline          \cite{mpc_mc} $P_{t_n}=20, update\_range= 5m$                        & 0.58 & 0.397 & 93.5  &   154.67  &  68.45 \\ \hline                      
\cite{mpc_mc} $P_{t_n}=40, update\_range= 5m$                        & 0.66  & 0.441 &  98.91  &  201.56   & 78.23 \\ \hline
\cite{tordesillas2019faster} $N_{whole} = N_{safe}=6$  $max\_poly = 3$ & 0.83  & $\approx 0.01$ &  54.56  &  81.45   & 49.39 \\ \hline
\cite{tordesillas2019faster} $N_{whole} = N_{safe}=12$ $max\_poly = 6$& \textbf{0.83}   & $ \approx \textbf{ 0.01} $  &  53.78  &  78.67   & 47.45 \\ \hline
Proposed $P_{t_n}=10, update\_range= 4m$ & 0.91    & $0.04 \pm 0.01$  &  56.78  &  68.78   & 49.86 \\ \hline
Proposed $P_{t_n}=15, update\_range= 6m$ & \textbf{1.0}   & $\mathbf{0.03}\pm \textbf{0.01}$ &   54.89  &  66.80   & 48.67 \\ \hline
\end{tabular}
\begin{tablenotes}
     \item[1]   $N_{whole}, N_{safe}$ : number of discretization points in the whole and safe trajectory,  $max\_poly$: maximum number of polydrons to represent the free space, $P_{t_n}$: NMPC prediction horizon length 
\end{tablenotes}
\end{table*}

\begin{figure}[ht!]
    \centering
\includegraphics[width=1\linewidth]{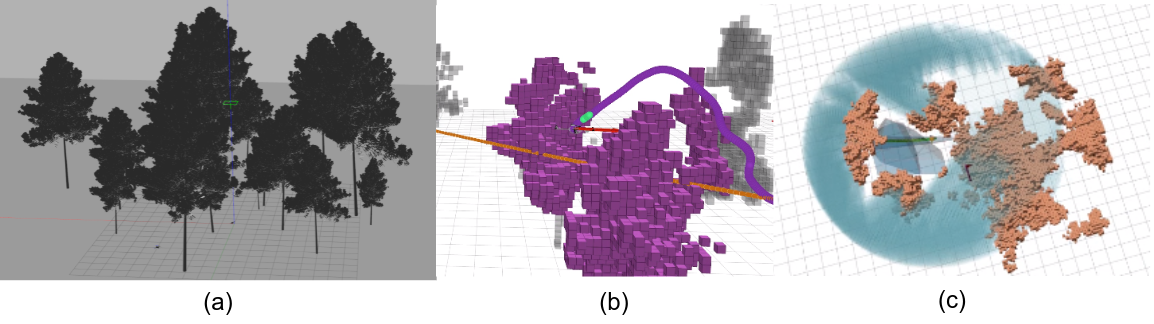}
    \caption{An example of testing the proposed (b) and FASTER (c) algorithms on a randomly generated forest (a)}
    \label{fig:testing_1}
    \vspace{-0.06in}
\end{figure}

In the final experiment, we have generated 12 random forests, e.g., Fig.~\ref{fig:testing_1}(a), where density (40m$\times$40m$\times$10m) was kept the same for all the environments. The three other methods: RRT*~\cite{kulathunga2020real}, a local planner~\cite{mpc_mc}, and FASTER~\cite{tordesillas2019faster} were used to validate the proposed approach. The results are provided in Table~\ref{t:comparision} and an example test case is shown in Fig.~\ref{fig:testing_1}. When the environment is cluttered, FASTER failed mainly due to the inability to find a path to local goal pose using JPS~\cite{zhou2017local}. In the proposed approach, the dead zone recovery technique tries to recover when the global planner fails to refine the trajectory and the local planner is also capable of planning ahead independently from the global planner. In consequence, the proposed approach has a higher success rate (number of times successfully reach the goal) compared to the other methods despite mean computation time (MCT) (ratio of total execution time to the total number of iterations) is slightly lower than FASTER. Since we are targeting low-speed maneuver, MCT is also acceptable. In each environment, the same start and goal poses were considered and the distance between them was set to 38.6m, ensuring no obstacle presence on those poses. There is no distinctive difference in the mean distance estimation between the proposed and FASTER.

\section{CONCLUSION}
This work presents a reference trajectories tracking approach for low-speed agile flights ensuring safety and dynamic feasibility in completely unknown environments. The essential properties of the proposed approach are online trajectory refinement and near-optimal control policy generation in parallel in horizon-based fashion, while only reasoning the surrounding environment. The proposed approach was tested on various simulated and real-world environments, achieving long range trajectory tracking. The local and global planners have mean computation times of approximately 0.06s  (15Hz)  and  0.05s  (20Hz), respectively, provided that tracking accuracy is less than 1m  in obstacle-free zones. We expect to extend this work for high-speed maneuvers in which we are going to focus on improving the local planner. The source code and complete experiments are available at Github\footnote[4]{\label{sourcecode} The source code and complete experiments -  \url{https://github.com/GPrathap/trajectory-tracker.git}}   


\section*{ACKNOWLEDGMENT}
This research has been financially supported by The Analytical Center for the Government of the Russian Federation (Agreement No. 70-2021-00143 dd. 01.11.2021, IGK 000000D730321P5Q0002))
\newpage
\bibliographystyle{IEEEtran}
\bibliography{references}


 





\end{document}